# Using Domain Knowledge for Low Resource Named Entity Recognition

Yuan Shi

School of Computer Science & Technology,
Beijing Institute of Technology, Beijing, China

## Abstract

*In recent years, named entity recognition has always been a popular research in the field of natural language processing, while traditional deep learning methods require a large amount of labeled data for model training, which makes them not suitable for areas where labeling resources are scarce. In addition, the existing cross-domain knowledge transfer methods need to adjust the entity labels for different fields, so as to increase the training cost.*

*To solve these problems, enlightened by a processing method of Chinese named entity recognition, we propose to use domain knowledge to improve the performance of named entity recognition in areas with low resources. The domain knowledge mainly applied by us is domain dictionary and domain labeled data. We use dictionary information for each word to strengthen its word embedding and domain labeled data to reinforce the recognition effect. The proposed model avoids large-scale data adjustments in different domains while handling named entities recognition with low resources. Experiments demonstrate the effectiveness of our method, which has achieved impressive results on the data set in the field of scientific and technological equipment, and the F1 score has been significantly improved compared with many other baseline methods.*

## Keywords

*Named Entity Recognition, Domain Knowledge, Low Resource, Domain Dictionary.*

## 1. Introduction

Named Entity Recognition (NER), also known as entity extraction technology, is one of the key technologies of natural language processing. Its goal is to identify the predefined entity categories from unstructured text, such as person, location, organization, etc. NER is conventionally formulated as a sequence labeling problem, mainstream method which has been widely used is neural network model[1] and pretrained Language Models (LMs)[2], for example, we often use Long Short Term Memory(LSTM)[3] or Bidirectional Encoder Representation from Transformers(BERT)[4] as encoding layer, multi-layer perceptron plus Softmax[5] or Conditional Random Field (CRF)[6] as decoding layer.

Supervised neural network model usually needs to use a large amount of labeled data for training. However, except for some mature fields, such as newswire domain, most other fields lack sufficient high-quality annotated corpus. When the annotated corpus is small, the existing method of entity recognition using deep neural network will have significant performance degradation. Due to the lack of sufficient training data, these models are unable to fully learn the implicit feature representation. It can be seen that named entity recognition in the field of resource scarcity is more difficult than that in the mature field.

In recent years, in order to solve the problem of named entity recognition in the field of resource scarcity, many methods have been proposed. Li used a large parallel corpus to project information from high resource domain to low resource domain[7]. Yang use cross-resource word vector to bridge low resource domain and high resource domain to realize knowledge transfer[8]. Chen and Ding propose explicitly connecting entity mentions based on both global coreference relations and local dependency relations for building better entity mention representations [9], which is desirable for domain-specific NER.

In our opinion, the problem of cross-domain knowledge transfer is that there are different entity categories in different fields. For example, in the newswire domain, we usually identify persons and locations, while in the science and technology equipment domain, we need to identify products and achievements. Since the decoding layer needs to be trained and tested with a consistent set of tags, the need to adjust the output layer to retrain in the new field may further increase training costs.

Based on this situation, we propose the use of domain knowledge to improve the problem of named entity recognition in resource scarcity domains. Specifically, analogy to the method of using word embedding to strengthen character embedding in Chinese NER, we introduce domain dictionaries and small-batch domain labeling data into this scenario, which effectively alleviates the NER problem in low resource domains. The feasibility of adopting this method is that although there is a lack of mature annotation data in many fields of scarce resources, some experience-generated domain dictionaries with rich information can be applied. These domain dictionaries contain a large number of entity information, which can effectively improve the recognition rate of domain-specific words. Our method is to scan the input sentence, for every word, find out whether it is located in a phrase in the domain dictionary. The dictionary vector is obtained by using the formula defined in subsequent chapters, and then spliced with pre-trained word embedding to obtain the enhanced word embedding. Small-batch domain labeling data are generated by concentrating a small number of professionals in a short period of time, we use it to replace generic domain data used by existing models as training set to achieve better recognition results. In this way, when dealing with different fields, only the dictionary and training data in this field need to be replaced. The structure of existing model itself needs no modification, which saves a lot of training costs.

In section 2, we introduce you some technical background. In section 3, we describe the principle and implementation of our model in detail, we conduct our experiments on the low resource domain dataset in section 4, respectively. The results show that our method has achieved very competitive results on the resource-scarce domain dataset of scientific and technological equipment, and the F1 score is significantly improved compared with many baseline methods. Finally, we summarize our article in section 5.

## 2. BACKGROUND

Generally speaking, Named Entity Recognition (NER) is to extract entities from unstructured texts. Academically, named entities generally include three categories (entities, time expressions and numerical expressions) and seven subcategories (persons, locations, organizations, time, date, currency, and percentage). In specific work, more categories of entities can be defined according to requirements. The most common modeling method is to model named entity recognition as a sequence labeling problem, that is, for an input token sequence, after model processing, the corresponding label will be assigned to each token as the output.

Domain dictionaries are lists of named entities collected from various sources, such as disease name dictionary in the medical field and actor name dictionary in the film field. They have been widely used in named entity recognition. Domain dictionaries are generally used to create features for the model, usually binary features, to indicate whether the corresponding n-gram exists in the dictionary. In addition, grammar features can be extracted from it, such as prefix and suffix, for rule matching.

Domain labeled data are necessary in the identification of named entities using neural network models. It will guide the model to learn implicit feature representation as training data, so that it can obtain the ability of identifying specific entities. In practical applications, due to the lack of mature annotation data in the field of resource scarcity, most of the existing methods use high-quality annotation corpus of general fields (such as newswire domain) as training dataset.

## 3. USING DOMAIN KNOWLEDGE

The basic model of our paper is LM-LSTM-CRF, namely, the 'Language Model–Long Short-Term Memory–Conditional Random Field 'model. Based on this model, domain knowledge is added as an auxiliary to improve the effect of the model in identifying named entities in the field of resource scarcity.

### 3.1. LM-LSTM-CRF

LM-LSTM-CRF was proposed by Liu et al., 2018[10], its main purpose is to solve the problem of lacking annotated corpus for model training in sequence annotation tasks. This model can be used for part-of-speech tagging, named entity recognition, noun phrase segmentation and other tasks that can be modeled as sequence annotation problems.

The model extracts knowledge from original text and assigns weight to the sequence annotation task. In pre-trained word vector, in addition to word level knowledge, character-aware neural language model is also introduced to extract character level knowledge. According to the experiment, when the tasks are inconsistent, if simply set them in a common training environment, the language model may affect the effect of sequence labeling. Therefore, the highway[11] layer has been used to convert the output of character level to different semantic spaces to mediate and unify these two tasks, and guide the language model to master the knowledge of specific tasks. Different from most transfer learning methods, this framework does not rely on any other supervision, but extracts knowledge from the self-contained order information of the training sequence. Compared with previous methods, this task-specific knowledge can be trained more effectively using simpler models. A large number of experiments on the benchmark data set has been conducted to prove the effectiveness of using character-level knowledge and collaborative training in sequence annotation tasks.

Overall, this model uses the character-level bidirectional LSTM structure to implement the language model, and forms a multi-task transfer learning layer with the sequence annotation feature extraction layer implemented by the word-level bidirectional LSTM structure. Finally, the CRF layer is used for sequence decoding.

### 3.2. Incorporating Domain Knowledge into Model

#### 3.2.1. Lattice-LSTM and Its Strengthening

In order to achieve the goal of integrating domain dictionary information into existing model, the method we apply is derived from Peng and Ma[12]. The main purpose of their work is to improve the Lattice-LSTM model.

The Lattice-LSTM model was proposed by Zhang and Yang[13]. For Chinese named entity recognition tasks, the model encodes input characters and all potential word sequences matched with dictionaries. Compared with previous character-based methods, this model can make better use of word and word sequence information. Compared with word-based methods, it can avoid named entity recognition errors caused by word segmentation errors. At the same time, due to the use of gated recurrent unit, the model can select the most relevant words from sentences to obtain better named entity recognition results.

Although the model has achieved some success in the task of Chinese named entity recognition, it still has a problem that cannot be ignored, that is, the model architecture is too complex. In essence, the model converts the input form of a sentence from a common chain sequence to a graph, which increases the computational cost of sentence modeling. It is this problem that limits its further application in the industrial field that requires real-time response.

Through the introduction of Lattice-LSTM, it can be seen that the advantage of Lattice-LSTM is to retain all possible matching words in the dictionary for each character, which can avoid the exploratory selection of character matching results and introduce error propagation. Therefore, if we want to optimize the model, we should retain this advantage as much as possible, meanwhile, try to overcome the problem of excessive modeling and computing costs caused by the conversion of input forms. In response to this, Peng and Ma proposed an improved method in [12].

The first method is based on Softword technology, which was originally used to merge word segmentation information into downstream tasks (Zhao and Kit, 2008[14]. Peng and Dredze, 2016[15]), mainly by embedding the corresponding segment labels to enhance character representation:

$$X_j^c \leftarrow \left[ X_j^c; e^{seg}(seg(c_j)) \right]. \qquad (1)$$

Here $X_j^c$ represents the character level vector, and $seg(c_j) \in y_{seg}$ denotes the segmentation label of the character $c_j$ predicted by the word splitter, $e^{seg}$ denotes the segmentation label embedding lookup table. Usually, $y_{seg} = \{B, M, E, S\}$, where B, M, E denote that the character is the beginning, middle, end of the word, and S denotes that the character itself constitutes a single word.

Based on this technology, a dictionary has been proposed to construct a word splitter, which allows a character to have multiple segmentation labels at the same time:

$$X_j^c \leftarrow [X_j^c; e^{seg}(segs(s)_j)]. \qquad (2)$$

Here $e^{seg}(segs(s)_j)$ is a 5-dimensional binary vector, and each dimension corresponds to each item of {B, M, E, S, O}. This method is called ExSoftword.

On the surface, ExSoftword has successfully introduced dictionary information into vector representation. However, through analysis, it can be found that although ExSoftword attempts to save all dictionary matching results by allowing a single character to have multiple segmented labels, it still loses a lot of information, so that in many cases it cannot recover the matching results from the segmentation label sequence.

In order to solve this problem, not only the possible segmented label characters, but also the corresponding matching words should be retained. The specific approach is to correspond each character c of a sentence s to four word sets marked by four segmentation labels ' BMES '. The word set B (c) consists of all dictionary matches beginning with c on $S$. Similarly, M(c) consists of all lexicon matched words in the middle of which c occurs, E(c) is composed of all dictionary matching words ending with c, and S(c) is a word composed of c. At this point, if the word set is empty, add a special word ' NONE '. Next, compress the four word sets of each character into a fixed dimension vector. To retain information as much as possible, four word sets are concatenated as a whole and added to the character representation:

$$e^s(B,M,E,S) = [v^s(B) \oplus v^s(M) \oplus v^s(E) \oplus v^s(S)],$$
$$X^c \leftarrow [X^c; e^s(B,M,E,S)]. \qquad (3)$$

Here, $v^s$ represents the function of mapping a single word set to a dense vector. This also means that each word set should be mapped to a fixed dimension vector. In order to achieve this goal, after trying the less successful mean-pooling algorithm, the word frequency has been used to represent its weight. Specifically, $w_c$ is used to represent the character sequence of $w$, and $z(w)$ is used to represent the frequency of $w_c$ in the statistical data set. Note that if $w_c$ is covered by another word in the dictionary, the frequency of w is not counted. Finally, the weighted representation of the word set $S$ is obtained by:

$$v^s(S) = \frac{1}{Z}\sum_{w \in S}(z(w)+c)e^w(w), \qquad (4)$$

Where

$$Z = \sum_{w \in B \cup M \cup E \cup S} z(w) + c.$$

Here, the weights of all words in the four word sets are normalized to allow them compete with each other across sets. $e^w$ indicates that the words are embedded in the query table. The addition of constant c is to introduce smoothing treatment to the weight of each word to increase the weight of rare words. Here, the value of c is set as: in the statistical data set, 10% of the training words appear less than c times.

To sum up, the word vector representation combined with dictionary information mainly includes the following four steps. First, a dictionary is used to scan each input sentence, and four word sets of ' BMES ' are obtained for each character in the sentence. Second, query the frequency of each word appearing on the statistics set. Third, get the vector representation of four word sets for each character by formula, and add it to the character representation. Finally, based on the enhanced character representation, any appropriate neural sequence labeling model can be used for sequence labeling.

### 3.2.2. Using Domain Dictionary

In our experiment, the experimental data we use is English text from science and technology equipment domain, however, the word vector representation combined with dictionary information introduced above is carried out on Chinese texts. If we want to apply this method in our own model, we need to redesign it to realize the conversion from Chinese processing to English processing.

Due to the language differences between Chinese and English, there is some distinction on the definition of character level in the two languages. When dealing with Chinese, Chinese characters are used as the basic character units. Depending on the difference of word segmentation, words composed of single and multiple characters can be used as word level units. Relatively speaking, in English, letters are used as basic character units and words are used as word level units. The main reason for this is that the characters in Chinese have the ideographic function, while in English only words have practical meaning, a single letter has no meaning. The method mentioned above is implemented for Chinese. The existing dictionary is used for matching to enhance the representation of character-level vector (for Chinese, that is, a single word). When dealing with English text, the existing English dictionary in the field of science and technology equipment (including a single word or phrases of multiple words) has been used by us to enhance the representation of word-level vector.

According to the differences between English and Chinese , our main processing flow is as follows:

(1). Input English sentence $S$. For each word $X$, find out whether it is located in a phrase in the domain dictionary, and process all the matching results into four word sets corresponding to BMES. These word sequences contain all the word sequences of the current word. For example, for the word 'rocket', if there are 'rocket motor ', ' multistage rocket ' and ' meteorological rocket ' in the dictionary, then B (rocket) = {{rocket motor}}, M (rocket) = {None}, E (rocket) = {{multistage rocket}, {meteorological rocket}}, S (rocket) = {None}.

(2). For the further apply, the word sequence set obtained in the previous step need to be compressed into a fixed-dimension vector, as shown in Formula (3), $v^s$ is the function of mapping the word set to a dense vector. Its definition is shown in Formula (4), $z(w)$ uses the frequency of the word sequence to represent the weight. We combine training and testing data of the task to construct a statistical data set, and conduct frequency statistics on it. Finally, the fixed-dimension vector is obtained through Formula (4).

(3). We add obtained vectors into the word representation. In the existing LM-LSTM-CRF model, the input used at the word level is Glove 100-dimension data, and the pre-trained word embedding is fine-tuned. In order to make use of the domain dictionary information, we use Glove 100-dimension data and the word vector augmented by domain dictionary information as input at the word level.

### 3.2.3. Adding Labeled Data

In the field of resource scarcity, in addition to the dictionary, there are some unlabeled data that can be used. These data belong to the original corpus, which can be used as testing data. We annotate some original corpus of science and technology equipment domain, and use these labeled data to replace the CoNLL-2003[16] set as a training set to retrain the model. The results show that higher F1 score is obtained by using the new model for experiments.

The text form of the original corpus is as follows:

*Launching the sixth branch of the US armed forces Defense News*
*Space*
*Launching the sixth branch of the US armed forces*
*By：Mike Rogers*

*Vandenberg Air Force Base supported the successful launch of the fourth Iridium mission on a SpaceX Falcon 9 rocket on Dec. 22，2017. (Tech. Sgt. Jim Araos/U.S. Air Force)*

The labeling method used in our work is BIO tagging, and the final labeling result is a combination of two types of labels. The first type of label indicates that the current entity belongs to the category defined from the meaning of entity itself, entity categories defined in this annotation are PER (persons), ORG (organizations), PCT (products), OUT (Outcomes), SER (Services), TIM (time). The second type of label indicates the position of the current word in the named entity, the standard practice is to use BIO labeling, that is, each word is labeled as ' B-X ', ' I-X ' or ' O ', and the ' X ' is the first type of label indicating that the current word belongs to type X, and these labels indicate that the current word is located at the beginning, the middle or end position of the named entity and does not belong to any named entity.

As for the annotation management, we use the method of cross-examination annotation by multiple annotators, that is, each annotator independently labels the data of his assigned part, and each document will be labeled by at least three annotators. Then each annotator reviews the work of other annotators to form a review chain, which promotes the communication between annotators and the unity of data understanding. The final annotation results of the data are obtained by voting of different annotators.

The final labeled text is as follows:

*Launching O*
*the O*
*sixth O*
*branch O*
*of O*
*the O*
*US B-ORG*
*armed I-ORG*
*forces I-ORG*
*Defense I-ORG*
*News O*

## 4. EXPERIMENT

In this section, we compare our method with existing baseline methods and demonstrate the effectiveness of our method by experimental results.

### 4.1. Experimental Setup

On the NER basic dataset, we use CoNLL-2003 dataset, which mainly contains four types of labels, namely PER (persons), LOC (locations), ORG (organizations), MISC (miscellaneous entities). The text in science and technology equipment domain is used as the dataset in the field of resource scarcity, and the characteristics of the text in this field are comprehensively analyzed. As mentioned in the preceding section, we mainly annotate six entities, namely PER (persons), ORG (organizations), PCT (products), OUT (Outcomes), SER (Services), TIM (time), etc. The experimental evaluation index is the F1 score of named entity recognition.

In terms of network structure setting, the standard of LM-LSTM-CRF model is mainly followed. The hyper-parameters of character-level LSTM and word-level LSTM are the same. The size of hidden layer is set to 300, the dimension of character-level embedding layer is 30, the dimension of word-level embedding layer is 100, and the depth of highway layer is set to 1.

As for the training optimization, the small-batch stochastic gradient descent method combined with momentum is used. The batch size and momentum are set to 10 and 0.9, respectively. The learning rate is set to $\eta_t = \frac{\eta_0}{1+\rho t}$, where $\eta_0$ is the initial learning rate and $\rho=0.05$ is the decay ratio.

In order to alleviate the over-fitting problem, the proportion of Dropout is fixed to 0.5. In order to resolve the gradient explosion problem that may occur in the training process, the gradient clipping strategy is adopted, and the threshold of gradient clipping is set to 5.0, when the gradient is updated, if the gradient vector is greater than this threshold, it will be limited in this range.

## 4.2. Results and Analysis

Our experiments are mainly compared with two baselines, the LM-BiLSTM-CRF model as the experimental basis, "LBC" for short, and the BERT-LSTM-CRF model with good performance in the named entity recognition task in recent years, "BBC" for short.

(1) Adding domain dictionary

First, we run the experiment of adding a domain dictionary, we call the LM-BiLSTM-CRF model with dictionary as Dicstrengthen-LM-BiLSTM-CRF model, "DLBC" for short. The training data used here comes from the Conll2003 data set. When training the DLBC model, use the dictionary of science and technology equipment mentioned before, and integrate its information into the word vector representation to obtain the target model. We use LBC model, BBC model and DLBC model for our experiments on the text in the field of scientific and technological equipment, and the results are shown in Table 1.

Table 1. F1-score (%) of three model

| Model | PER | LOC | ORG | OVERALL |
|-------|-----|-----|-----|---------|
| LBC | 80.32 | 81.25 | 79.57 | 80.38 |
| BBC | 83.88 | 82.54 | 83.06 | 83.16 |
| DLBC | 85.72 | 81.53 | 85.21 | 84.15 |

It can be seen from the results in the table that the F1 value of each model is generally not high, which is mainly because the training data used by these models are from the Conll2003 data set, and the field characteristics mastered in the training are not suitable for the field of scientific and technological equipment. Comparing the results of each model, it can be seen that the DLBC model with dictionary has achieved the best results in the name entity PER, the organization entity ORG and overall F1, especially compared with the LBC model without dictionary. However, in terms of location entity recognition, the DLBC model is less effective than the BBC model, basically similar to the LBC model, mainly because of the low priority of location entities in the field of science and technology equipment, and the dictionary we collect does not contain many location entities, so the results are basically the same as when no domain dictionary is added. Overall, adding domain dictionaries can effectively improve the recognition effect of named entity recognition model.

(2) Adding small batch labeled data

Next, we test the effect of adding small batch labeled data on the experimental results. The LM-BiLSTM-CRF model using small batch labeled data is called LM-BiLSTM-CRF-annplus,

"LBC+" for short. The small batch labeled data used are provided by the concentrated annotation of some experts in a short time. The original texts are news texts from the field of science and technology equipment. This experiment will explore the relationship between the size of labeled text and model results, as shown in Figure 1 and Figure 2.

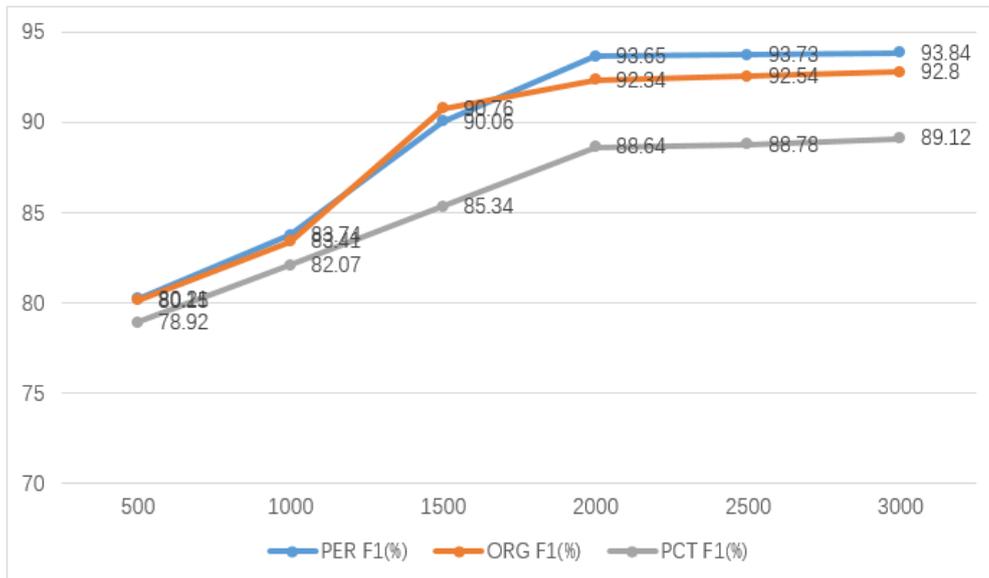

Figure 1. Named Entity Recognition Results Varying with Data Scale 1

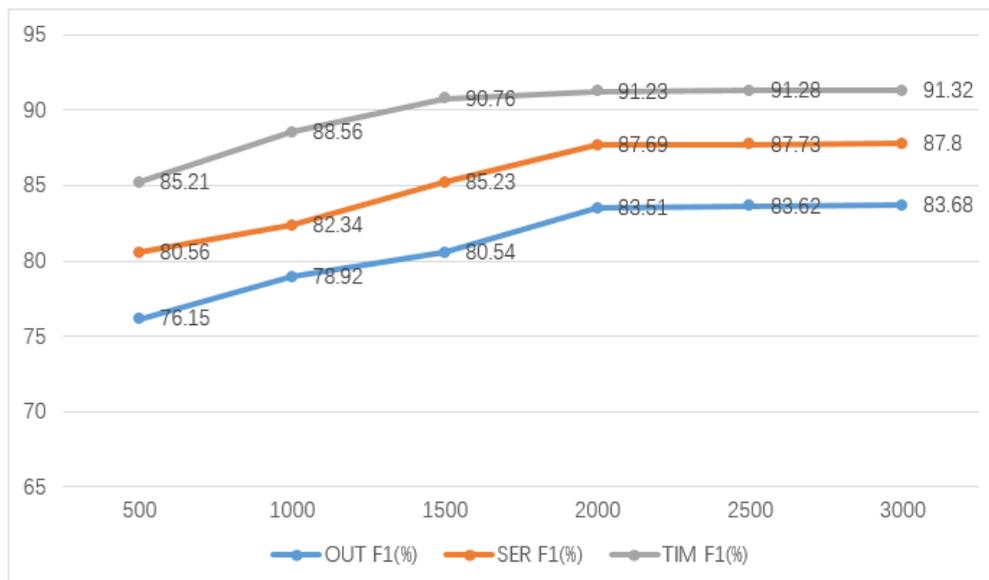

Figure 2. Named Entity Recognition Results Varying with Data Scale 2

These two figures show the change of entity recognition results with the scale of labeled data. Here we define six entities, namely PER (name), ORG (organization), PCT (product), OUT (result), SER (position) and TIM (time). It can be seen from the figure that at the beginning, when the scale of labeled documents gradually increases, the results of entity recognition also increase. However, when the scale of labeled documents reaches more than 2000, even if the labeled documents continue to increase, the F1 value basically no longer increase or only slightly increase. Considering the increase of annotation cost, we finally decide that the appropriate scale of labeled documents should be set at 2000.

In addition, it can be seen from Figure 2 that compared with other types of entities, the recognition results of TIM (time) do not improve significantly with the increase of the scale of labeled data, which is mainly because the form of time entities in the text is relatively fixed. Even if we increase the scale of labeled documents, the model cannot obtain more time entity features.

(3) Adding domain dictionary and small batch labeled data

Finally, we add the domain dictionary and small batch annotation data to the model at the same time, and obtain our final model Dicstrength-LM-BiLSTM-CRF-annplus, "DLBC+" for short. The experimental results of this model compared with other baseline models are shown in Table 2.

Table 2. F1-score (%) of three model

| Model | PER | ORG | PCT | OUT | SER | TIM | OVERALL |
|---|---|---|---|---|---|---|---|
| LBC+ | 93.65 | 92.34 | 88.64 | 83.51 | 87.69 | 91.23 | 89.51 |
| BBC+ | 94.15 | 92.89 | 90.71 | 85.29 | 89.36 | 95.08 | 91.25 |
| DLBC+ | 96.24 | 94.37 | 91.75 | 87.65 | 88.14 | 96.76 | 93.05 |

Based on the above experimental results, it can be seen that the LM-LSTM-CRF model combined with the dictionary is better than LM-LSTM-CRF model and BERT-LSTM-CRF model on the whole, because the domain-specific name information contained in the dictionary improves the recognition rate in general, word embedding with dictionary enhancement carries more useful information for recognition. Then, the model using domain labeled data as training data performs significantly better than the model using CoNLL-2003 dataset as training data in terms of person and organization, which is also well understood, because the domain information contained in the domain data is much more than that contained in the general data set, through the study of the common information in the field, the model grasps more domain-specific implicit features and so as to obtains better results.

In addition, compared with the improvement of 'DLBC+' on 'LBC+', the improvement of 'DLBC' on 'LBC' is more obvious. This is because when the domain dictionary is applied to the model with domain labeled data as the training set, some domain proper names have been labeled, that is, some information in the dictionary has been used, so the improvement effect is relatively less obvious. Finally, ' BBC+' is superior to ' DLBC+' in the recognition of SER (service), which may be due to the fact that some ambiguous words in the dictionary disturb the experimental results, and also represent that the quality of the current dictionary can be improved.

Although the experimental results show that our model can solve the problem of named entity recognition in many low resource domains, there are still some limitations and deficiencies. It depends on the quality of the domain dictionary, if there is a lack of high-quality domain dictionaries in the field to be processed, then we will need to invest more costs in domain data annotation to achieve better processing results, thereby increasing manpower consumption. We plan to improve this problem in our further research in the future.

## 5. CONCLUSIONS

We propose a method to improve named entity recognition in low resource domain using domain knowledge, and present its effectiveness in the field of science and technology equipment. The experimental results show that our method has achieved competitive results in low resource domain, and F1 score has been significantly improved compared with other traditional baseline methods.

In addition, on the basis of the existing work, our next steps mainly include:

(1) Experiments in more resource scarce fields, and developing corresponding processing schemes for domain characteristics to improve the model processing effect.
(2) Apply the dictionary-based approach to more existing models to see if we can improve their performance.
(3) Conduct experiments on larger datasets to observe the improvement of model performance and explore the most appropriate training dataset size for our model.

## ACKNOWLEDGEMENTS

This research work is financially supported by The National Key Research and Development Program of China (NO.2017YFB1002101). This work was also funded by the Institute of Science and Development, Chinese Academy of Sciences (NO.GHJ-ZLZX-2020-42) and the Joint Advanced Research Foundation of China Electronics Technology Group Corporation (CETC) (No. 6141B08010102). The authors want to thank for the helpful suggestions from Heyan Huang, Chong Feng, Bo Wang and Ximo Bian. The corresponding author of this article is Yuan Shi.

## REFERENCES

[1] Ma, X., & Hovy, E. (2016). End-to-end sequence labeling via bi-directional lstm-cnns-crf. arXiv preprint arXiv:1603.01354.
[2] Akbik, A., Blythe, D., & Vollgraf, R. (2018, August). Contextual string embeddings for sequence labeling. In Proceedings of the 27th international conference on computational linguistics (pp. 1638-1649).
[3] Hochreiter, S., & Schmidhuber, J. (1997). Long short-term memory. Neural computation, 9(8), 1735-1780.
[4] Devlin, J., Chang, M. W., Lee, K., & Toutanova, K. (2018). Bert: Pre-training of deep bidirectional transformers for language understanding. arXiv preprint arXiv:1810.04805.
[5] Strubell, E., Verga, P., Belanger, D., & McCallum, A. (2017). Fast and accurate entity recognition with iterated dilated convolutions. arXiv preprint arXiv:1702.02098.
[6] Lafferty, J., McCallum, A., & Pereira, F. C. (2001). Conditional random fields: Probabilistic models for segmenting and labeling sequence data.
[7] Li, Q., Li, H., Ji, H., Wang, W., Zheng, J., & Huang, F. (2012, October). Joint bilingual name tagging for parallel corpora. In Proceedings of the 21st ACM international conference on Information and knowledge management (pp. 1727-1731).
[8] Yang, Z., Salakhutdinov, R., & Cohen, W. W. (2017). Transfer learning for sequence tagging with hierarchical recurrent networks. arXiv preprint arXiv:1703.06345.
[9] Chen, P., Ding, H., Araki, J., & Huang, R. (2021, January). Explicitly Capturing Relations between Entity Mentions via Graph Neural Networks for Domain-specific Named Entity Recognition. In Proceedings of the 59th Annual Meeting of the Association for Computational Linguistics and the 11th International Joint Conference on Natural Language Processing (Volume 2: Short Papers).
[10] Liu, L., Shang, J., Ren, X., Xu, F., Gui, H., Peng, J., & Han, J. (2018, April). Empower sequence labeling with task-aware neural language model. In Proceedings of the AAAI Conference on Artificial Intelligence (Vol. 32, No. 1).

[11] Srivastava, R. K., Greff, K., & Schmidhuber, J. (2015). Highway networks. arXiv preprint arXiv:1505.00387.
[12] Ma, R., Peng, M., Zhang, Q., & Huang, X. (2019). Simplify the usage of lexicon in Chinese NER. arXiv preprint arXiv:1908.05969.
[13] Zhang, Y., & Yang, J. (2018). Chinese NER using lattice LSTM. arXiv preprint arXiv:1805.02023.
[14] Zhao, H., & Kit, C. (2008). Unsupervised segmentation helps supervised learning of character tagging for word segmentation and named entity recognition. In Proceedings of the Sixth SIGHAN Workshop on Chinese Language Processing.
[15] Peng, N., & Dredze, M. (2016). Improving named entity recognition for chinese social media with word segmentation representation learning. arXiv preprint arXiv:1603.00786.
[16] S Sang, E. F., & De Meulder, F. (2003). Introduction to the CoNLL-2003 shared task: Language-independent named entity recognition. arXiv preprint cs/0306050.

## Author

**Shi Yuan**, born in 1995, master of science in cyberspace security at School of Computer Science & Technology Beijing Institute of Technology. His main research interest is named entity recognition. (18810932070@163.com)

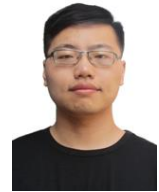